\documentclass[letterpaper, 10 pt, conference]{ieeeconf}  
\IEEEoverridecommandlockouts                              

\usepackage[ansinew]{inputenc}
\usepackage{amsmath}
\usepackage{amssymb}
\usepackage{amsfonts}
\usepackage{graphicx}
\usepackage{color}
\usepackage{array}
\usepackage{floatflt}
\usepackage{float}
\usepackage{multirow}
\usepackage{math}
\usepackage{dsfont}
\usepackage{relsize}
\usepackage{cite}
\usepackage[table]{xcolor}
\usepackage{balance}

\renewcommand{\vec}[1]{\boldsymbol{#1}}

\newcommand{\wrt}{w.\,r.\,t.\ }

\newcommand{\eg}{e.\,g.}
\newcommand{\q}{\vec{q}}
\newcommand{\dq}{\dot{\vec{q}}}
\newcommand{\ddq}{\ddot{\vec{q}}}
\newcommand{\thetab}{\vec{\theta}}
\newcommand{\dth}{\dot{\vec{\theta}}}
\newcommand{\ddth}{\ddot{\vec{\theta}}}
\newcommand{\M}{\vec{M}}
\newcommand{\B}{\vec{B}}
\newcommand{\K}{\vec{K}}
\newcommand{\D}{\vec{D}}

\newcommand{\sect}[1]{{Section~\ref{#1}}}
\newcommand{\fig}[1]{{Fig.~\ref{#1}}}
\newcommand{\oB}{\left(\begin{matrix}}
\newcommand{\cB}{\end{matrix}\right)}

\newcommand{\slow}{\text{slow}}
\newcommand{\fast}{\text{fast}}
\newcommand{\full}{\text{full}}

\title{\LARGE Model Predictive Control for Flexible Joint Robots}
\author{Maged Iskandar, Christiaan van Ommeren, Xuwei Wu, Alin Albu-Schäffer, and Alexander Dietrich
\thanks{The authors are with the Institute of Robotics and Mechatronics, German Aerospace Center (DLR), Wessling, Germany, {maged.iskandar@dlr.de}}}

\begin{document}
\maketitle
\thispagestyle{empty}
\pagestyle{empty}

\begin{abstract}
Modern Lightweight robots are constructed to be collaborative, which often results in a low structural stiffness compared to conventional rigid robots. 
Therefore, the controller must be able to handle the dynamic oscillatory effect mainly due to the intrinsic joint elasticity. 
Singular perturbation theory makes it possible to decompose the flexible joint dynamics into fast and slow subsystems. 
This model separation provides additional features to incorporate future knowledge of the joint-level dynamical behavior within the controller design using the Model Predictive Control (MPC) technique.  
In this study, different architectures are considered that combine the method of Singular Perturbation and MPC. 
For Singular Perturbation, the parameters that influence the validity of using this technique to control a flexible-joint robot are investigated. 
Furthermore, limits on the input constraints for the future trajectory are considered with MPC. 
The position control performance and robustness against external forces of each architecture are validated experimentally for a flexible joint robot.
\end{abstract}


\section{Introduction}\label{sec:introduction}
The lightweight robot design is a beneficial concept in various aspects such as energy efficiency, payload-to-weight ratio, and safety features in human-robot interaction \cite{2007.Schaeffer.DLR_LWR,Iskandar2021}.
One of the main challenges related to that is to design a controller that can handle both the oscillatory behavior (local joint vibrations), caused by the mechanical flexibility, and the motor constraints (\eg, torque or velocity limits). If both are not considered properly, instabilities can occur \cite{de2016robots}. Therefore, for model-based control, it is crucial to provide a sufficiently accurate model within the range of operation. 

Various approaches are known from the literature to control a robot with flexible elements \cite{2002.Ott.Comparison,1996.DeLuca.Tomei,2018.Kim,Iskandar2020a}. A common technique that aims at splitting the system dynamics into a slow and fast subsystem and enabling the design of the control law for both the slow and the fast models independently is Singular Perturbation (SP) \cite{khalil2002b}. The control law of the link-dynamics is usually obtained based on the approximate slow model of the zeroth order \cite{2002.Ott.Comparison}, and the joint-level torque control loop is designed and regulated separately within the SP approach. 
In addition to that, parameter uncertainties are considered through a regression vector. 
An extension of the control law by corrective terms is suggested in~\cite{1996.DeLuca.Tomei}, while in \cite{2018.Kim} the slow model of the flexible-joint robot is improved by adding an additional perturbation term which reduces the overall tracking error. Furthermore, in \cite{Iskandar2020a} shaping parts of the system dynamics using a desired dynamical behavior is achieved to control the flexible-joint in the link-side directly. As known from the literature, the preservation of the natural inertia is highly beneficial in terms of robustness \cite{DIETRICH2021104875}.

The main restrictions of SP are the inability to incorporate actuator saturation and the fact that only current reference points are considered. A technique that addresses these drawbacks is Model Predictive Control (MPC). It has been applied across different areas, such as autonomous driving \cite{kong2015kinematic}, drones \cite{2021.Dyro}, or interconnected tanks \cite{2020.Gonzalez}. Extensions of the standard linear or non-linear MPC involve passivity-based state constraints \cite{2007.Raff}, the addition of safety filters \cite{2021.Wabersich.PredictiveSafetyFilter}, the consideration of process uncertainty \cite{2019.Racovic.RMPC}, or various learning components \cite{2020.Hewing.LBMPC}. In the area of flexible joint robots, the authors in \cite{2014.Wei} have implemented an explicit MPC with the advantage that hardware-requirements for real-time execution are lower. However, any offline solutions need to be recalculated if the model is modified. In \cite{2019.Hasan} gravity is considered in the model linearization which results in an adaptive MPC, as it adapts its linear model according to the link position.
In general, the MPC approach can be combined with SP. For example, in \cite{2014.Sanilla} the SP is applied to separate into a slow and a fast model, then MPC is applied to control both of them. In \cite{2020.Fehr} a vision-based MPC approach is developed to handle the joint flexibility and dampen out the corresponding end-effector vibrations.
\begin{figure}[t]
	\centering
	\includegraphics[width=1\linewidth]{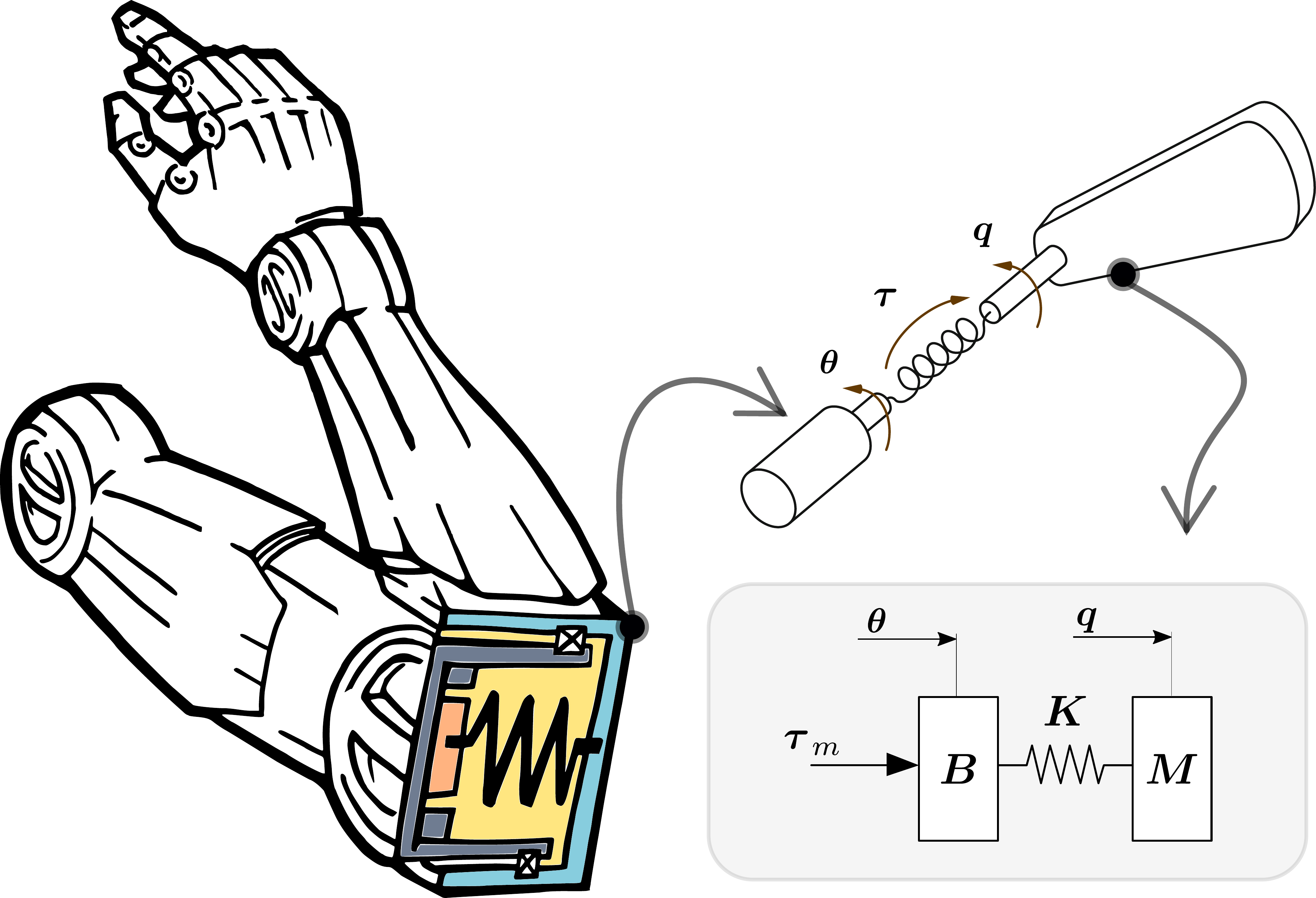}
	\caption{Conceptual example of flexible joint robot and its simplified single joint dynamical model. The elastic joint torque is transmitted between the inertia of the motor and the link via the joint stiffness.}
	\label{fig:intro_fig}
\end{figure}

One of the main challenges of applying SP to flexible-joint dynamics is to identify under which circumstances this theory can be applied. The reason is that an upper bound of the perturbation parameter can still exist, even though it is approximated to be zero in the slow model \cite{2006.Subudhi,2006.Karimi}. This upper bound can be found by the contraction mapping principle \cite{1999.Kokotovic.SingPert} or based on the frequency domain approach \cite{1990.Chen}. When a simplified model \cite{spong1989a} is considered, this parameter is typically proportional to the inverse of the square of the joint stiffness. Since the overall dynamics is altered by the control law, the relative distance towards the upper bound can be manipulated. However, even if this upper bound would be higher than the actual perturbation parameter, instabilities could still occur due to actuator saturation. Therefore, it is crucial that the parameters of the control law are aligned with the capabilities of the system to achieve higher control performance.

In this work, three different (multiple) MPC structures are proposed for the general flexible-joint robot dynamics. The control structures follow the concept of the two-time-scales of SP theory. 
The approach allows to impose constraints into the control law \eg\, the maximum allowable torque of the joints actuators.
Experimental comparisons on a flexible joint robot demonstrate the performance of the MPC framework, also \wrt the classical SP approach and a standard motor-PD controller. 
In addition, the validity of the SP method is investigated \wrt joint stiffness and actively shaping the motor inertia at different control frequencies. Furthermore, the approach allows to realize link-side impedance/position actions, enabling direct controllable interaction with the environment. 

The paper is organized as follows. In Section~\ref{sec:Fundamentals},  the system model to be used throughout this work is introduced, and an overview of the standard SP approach for the flexible joint robot is provided. The proposed MPC design idea and structures are presented in Section Section~\ref{sec:methods}. Experimental results and validations of the control design approach are shown and discussed in Section~\ref{sec:simulations_and_experiments}. Finally, the conclusion in Section~\ref{sec:conclusions} closes the paper.
\section{Fundamentals}\label{sec:Fundamentals}
The dynamical equations of flexible-joint robots require twice the number of generalized coordinates compared to rigid-body systems. For revolute-joint robots, they can be written based on the assumptions that the joint elasticity can be modeled through a linear torsional spring and the kinetic energy of each motor is only due to its own rotation \cite{spong1989a}:
\begin{align}
\M(\q) \ddq +\vec{C}(\q,\dq) \dq+ \vec{g}(\q) &= \K(\thetab-\q) +\vec{\tau}_{\text{ext}}\label{eq:link_dynamics} \ , \\
\B \ddth + \vec{K}(\thetab-\q)	&= \vec{\tau}_{\text{m}}   \label{eq:motor_dynamics} \ .
\end{align}
Herein, ${\q\in\mathbb{R}^{n}}$ represents the vector of the $n$ link-side coordinates and ${\thetab \in\mathbb{R}^{n}}$ describes the corresponding motor position coordinates. Let ${\vec{\tau}= \K(\thetab-\q)}$ be the elastic joint torque which is transmitted between motor and link, with the positive definite joint stiffness ${\K \succ 0}$. 
Gravitational torques are represented by ${\vec{g}(\q)\in\mathbb{R}^{n}}$, and the symmetric and positive definite link-side inertia matrix is defined by ${\M(\q)\in\mathbb{R}^{n\times n}}$. Also, ${\vec{C}(\q,\dq)\in\mathbb{R}^{n\times n}}$ is the Coriolis/centrifugal matrix which is formulated such that the skew-symmetry $\dot{\M}(\q) = \vec{C}(\q,\dq) +\vec{C}(\q,\dq)^T$ holds. External torques are described by $\vec{\tau}_{\mathrm{ext}}$.
The motor inertia is given by ${\B\in\mathbb{R}^{n\times n}}$ and accelerated by the motor torque $\vec{\tau}_{\text{m}} $.\footnote{The motor is modeled as an ideal torque source neglecting underlying (electrical) dynamics.}
Here, a wider class of systems can be considered without loss of generality \eg\, to include prismatic joints.

Commonly, the torque dynamics can be regulated by using the singular-perturbation (SP) approach \cite{khalil2002b,2008.Ott.CartImp} as     
\begin{equation}
\vec{\tau}_{\text{m}} = \vec{\tau}_\mathrm{d} + \K_\mathrm{T}(\vec{\tau}_\mathrm{d}-\vec{\tau})-\epsilon\K_\mathrm{S}\dot{\vec{\tau}} \ . \label{eq:SP_control}
\end{equation}
That can be interpreted as classical PD control for torque regulation \cite{albuschaeffer2007a,de2016robots}.
The desired joint torque $\vec{\tau}_\mathrm{d}$ is fed forward and the (measured) joint torque $\vec{\tau}$ and its time derivative $\dot{\vec{\tau}}$ are used in the feedback loop incorporating the corresponding positive definite gain matrices $\K_\mathrm{T}$, $\K_\mathrm{S}$, and $\epsilon$ is a small positive parameter. The term $\dot{\vec{\tau}}$ is usually derived numerically, based on $\vec{\tau}$, and $\epsilon$ is motivated by the singular-perturbation method and is responsible for the two time-scale separation. The dynamics \eqref{eq:link_dynamics}--\eqref{eq:motor_dynamics} can be rewritten as 
\begin{align}
\M(\q) \ddq +\vec{C}(\q,\dq) \dq+ \vec{g}(\q) &= \vec{\tau} +\vec{\tau}_{\text{ext}}\label{eq:link_dynamics_tau} \ , \\
\B \K^{-1} \ddot{\vec{\tau}} + \vec{\tau}	&= \vec{\tau}_{\text{m}}  -\B \ddq \label{eq:motor_dynamics_tau} \ .
\end{align}

According to SP theory the system can be expressed in a two-time-scale manner, with
\begin{equation}
\begin{split}  
(\M(\q_\slow)+\B) \ddq_\slow +\vec{C}(\q_\slow,\dq_\slow) \dq_\slow \\ + \vec{g}(\q_\slow)   =\vec{\tau}_{\text{m,}\slow} +\vec{\tau}_{\text{ext,}\slow} \\\label{eq:slow_dynamics}
\end{split}
\end{equation}
for the slow dynamics.
The so-called boundary layer system can be obtained through the deviation of the actual joint torque from its quasi-steady-state value. That also defines the fast component of the torque as
\begin{equation}
\label{eq:fast_total_torque}
\vec{\tau}_\fast = \vec{\tau} - \vec{\tau}_\slow\,,
\end{equation}    
where $\vec{\tau}_\slow$ is the slow component of the joint torque.
By applying the SP approach to the flexible-joint dynamics, one obtains two subsystems that can be separately controlled. Intuitively speaking, the slow part \eqref{eq:slow_dynamics} forms an equivalent rigid-body model, which one also obtains when neglecting the motor or torque dynamics. 
From that perspective the classical SP control law \eqref{eq:SP_control} can be straightforwardly reformulated as
\begin{equation}
\vec{\tau}_{\text{m}} = \underbrace{(\vec{I}+\K_\mathrm{T})\vec{\tau}_\mathrm{d}-\K_\mathrm{T}\vec{\tau}_\text{slow}}_{\mathrm{Slow\,component}} + \underbrace{\K_\mathrm{T}(\vec{\tau}_\text{slow}-\vec{\tau})-\epsilon\K_\mathrm{S}\dot{\vec{\tau}}}_{\mathrm{Fast\,component}}
\end{equation}
which results in the slow controlled link-side dynamics
\begin{equation}
(\M(\q)+(\vec{I}+\K_\mathrm{T})^{-1}\B) \ddq +\vec{C}(\q,\dq) \dq+ \vec{g}(\q) = \vec{\tau}_\mathrm{d} \ .
\end{equation}
Analogous to \eqref{eq:slow_dynamics}, it has the form of the rigid-body model with inertia matrix ${\M(\q)+(\vec{I}+\K_\mathrm{T})^{-1}\B}$. From a physical point of view, this can be intuitively interpreted as an active reduction of the motor inertia from $\B$ to the desired value $\B_\mathrm{d}$ when choosing $\K_\mathrm{T} = \B\B_\mathrm{d}^{-1}-\vec{I}$ in \eqref{eq:SP_control}. In this case the torque control loop can be specified through the damping parameter $\K_\mathrm{S}$ and the ratio of the reduction of the apparent motor inertia $\B\B_\mathrm{d}^{-1}$.
The commanded motor torque \eqref{eq:SP_control} can be combined with a link-side PD tracking controller
as
\begin{equation}
\vec{\tau}_\mathrm{d} = \vec{g}(\q) + (\M(\q)+\B_\mathrm{d}) \ddq_\mathrm{d} -\K_{q} (\q-\q_\mathrm{d})  - \D_{q} (\dq-\dq_\mathrm{d}), \label{eq:qposecont}
\end{equation}
with ${\K_{q}, \D_{q} \succ 0}$ being the desired link-side stiffness and damping, respectively, and $\q_\mathrm{d}$ is the desired link position. Noticeably, \eqref{eq:qposecont} is obtained using link-side coordinates but conventionally the motor-side position and velocity are used to generate the desired torque \cite{albuschaeffer2007a}. Interestingly, using the link-side coordinates, one can consider the classical control methods applied on rigid-body dynamics \cite{paden1988globally}.

In \eqref{eq:link_dynamics}--\eqref{eq:motor_dynamics}, dissipative friction effects can be included in different forms. The most dominant component in flexible-joint robots is motor-side friction due to gear friction \cite{albuschaeffer2007a}, which is actively reduced by the factor $\B\B_\mathrm{d}^{-1}$ as a result of \eqref{eq:SP_control}. Further friction effects can be considered by using model-based compensation techniques \cite{wolf2018extending,Iskandar2019a} or the use of motor-side friction observers \cite{le2008friction}. Nevertheless, this work focuses on the effects of joint elasticity, and therefore, friction effects are not explicitly considered in the following analysis.

\section{Methods}\label{sec:methods}
   Inspired by SP theory, the two-time-scale property can be used to apply the MPC technique in order to handle different ranges of dynamical effects. This also allows to impose practically motivated constraints within the low-level joint control to consider the physical limitations of the system actuation.
\subsection{MPC-Fast (motor dynamics)}
Since SP theory allows the separation between fast dynamics (torque dynamics) and slow dynamics (rigid-body link dynamics), it creates the prerequisites to apply MPC for the low-level torque control loop.
The joint stiffness $\K$ is related to the perturbation parameter $\epsilon$ through
\begin{equation}
\label{eq:sp_K_K0}
\K = \frac{\K_\epsilon}{\epsilon^2} \ .
\end{equation}
Let ${\K_\epsilon \succ 0}$ be a diagonal matrix. Using \eqref{eq:link_dynamics_tau}--\eqref{eq:motor_dynamics_tau} the dynamics can be reformulated\footnote{Dependencies on the states have been omitted for the sake of readability.} as
\begin{equation}
\label{eq:ddtau}
\epsilon^2\ddot{\vec{\tau}}  = \K_\epsilon(\B^{-1}\vec{\tau}_{\text{m}} - (\M^{-1} + \B^{-1})\vec{\tau} + \M^{-1}\vec{n}) \ ,
\end{equation}
with ${\vec{n} = \vec{C}(\q,\dq) + \vec{g}(\q)}$. Analogous to \eqref{eq:fast_total_torque}, the motor torque is composed of slow and fast terms as
\begin{equation}
\vec{\tau}_{\text{m}} = \vec{\tau}_{\text{m,}\slow} + \vec{\tau}_{\text{m,}\fast} 
\end{equation}
When ${\epsilon\rightarrow 0}$ in \eqref{eq:ddtau} the slow component of the torque dynamics $\vec{\tau}_\slow|_{\epsilon \rightarrow 0}$ can be obtained.
\begin{equation}
\label{tau_s}
\vec{\tau}_\slow = (\M^{-1} + \B^{-1})^{-1}(\B^{-1}\vec{\tau}_{\text{m,}\slow} + \M^{-1}\vec{n}) \ .
\end{equation}
Using \eqref{eq:fast_total_torque} the fast component of the torque dynamics is
\begin{equation}
\vec{\tau}_\fast = \vec{\tau} - (\M^{-1} + \B^{-1})^{-1}(\B^{-1}\vec{\tau}_{\text{m,}\slow} + \M^{-1}\vec{n}) \ .
\label{eq:tau_fast}
\end{equation}
By means of SP \cite{ott2002comparison} and through the substitution in \eqref{eq:ddtau} the fast time scale can be introduced as $\nu=t/\epsilon$ with time $t$. Thus, the second derivative of \eqref{eq:tau_fast} \wrt the time scale $\nu$ is\footnote{The first and second time derivatives in the normal time scale $t$ are ${\dot{\vec{z}}:= \frac{d\vec{z}}{dt}, \ddot{\vec{z}}:= \frac{d^2\vec{z}}{dt^2}}$, while in the fast time scale $\nu$, they are given by ${\vec{z}':= \frac{d\vec{z}}{d\nu}, \vec{z}'':= \frac{d^2\vec{z}}{d\nu^2}}$, with the relations ${\vec{z}'=\epsilon\dot{\vec{z}}, \vec{z}''=\epsilon^2\ddot{\vec{z}}}$.} 
\begin{equation}
{\vec{\tau}''_\fast} = \vec{\tau}'' = {\epsilon^2}\ddot{\vec{\tau}}
\end{equation}
substituting $\ddot{\vec{\tau}}$ from \eqref{eq:ddtau}. Keeping in mind that the dynamics can be expressed in the fast time scale, one obtains 
\begin{equation}
{\vec{\tau}''_\fast} = - \K_\epsilon (\M^{-1}+\B^{-1})\vec{\tau}_\fast + \K_\epsilon \B^{-1}\vec{\tau}_{\text{m,}\fast} \,.
\label{eq:tauf_fast}
\end{equation}
At this point \eqref{eq:tauf_fast} can be converted back to the normal time scale by $\vec{\tau}''_\fast=\epsilon^2 \ddot{\vec{\tau}}_\fast$, that is,
\begin{equation}
\ddot{\vec{\tau}}_\fast = -\K(\M^{-1}+\B^{-1}) \vec{\tau}_\fast + \K\B^{-1}\vec{\tau}_{\text{m,}\fast} \ .
\end{equation}
That yields the full representation of the fast torque dynamics featuring
\begin{equation}
\label{fast_mpc}
\begin{bmatrix} 
\dot{\vec{\tau}}_\fast\\ \ddot{\vec{\tau}}_\fast
\end{bmatrix}
\!\!=\!\!
\underbrace{
	\begin{bmatrix} 
	\vec{0} & \vec{1} \\ -\K(\M^{-1}\!+\!\B^{-1}) & \vec{0} 
	\end{bmatrix}}_{\displaystyle  \vec{A}_\fast}
	\begin{bmatrix} 
\vec{\tau}_\fast \\ \dot{\vec{\tau}}_\fast
\end{bmatrix}
\!+\!
\underbrace{  
\begin{bmatrix} 
\vec{0} \\ \K\B^{-1}
\end{bmatrix}}_{ \displaystyle \vec{E}_\fast}
\vec{\tau}_{\text{m,}\fast} \ .
\end{equation}
The model \eqref{fast_mpc} can be used as a reference to predict the behavior of the fast torque dynamics, which is directly related to local vibrations in the joints. Based on this prediction it is possible to determine the motor torque respecting physical constraints, including the maximum feasible motor torque. 
\subsection{MPC-Slow (link dynamics)} 
Similarly, the link-side dynamics \eqref{eq:slow_dynamics} can be targeted in terms of MPC. Then the outer loop is controlled that way while the inner torque loop is controlled through the classical SP scheme \eqref{eq:SP_control}. Thus, the MPC is responsible for the slow component in the commanded motor torque. Therefore, the slow dynamics is used to compute the link velocity and acceleration as following
\begin{equation}
\label{mpcslow}
\begin{bmatrix} 
	\dq\\ \ddq
\end{bmatrix}
\!\!=\!\!
\underbrace{
\begin{bmatrix} 
	\vec{0} & \vec{1} \\ \vec{0}  & -(\M+\B)^{-1}\vec{C}
\end{bmatrix}}_{\displaystyle \vec{A}_\slow}
\begin{bmatrix} 
	\q \\ \dq
\end{bmatrix}
+
\underbrace{ 
\begin{bmatrix} 
	\vec{0} \\ (\M+\B)^{-1}
\end{bmatrix}}_{\displaystyle \vec{E}_\slow}
\vec{\tau}_{\text{m,}\slow} \ .
\end{equation}
\begin{figure}[H]
	\centering
	\includegraphics[width=1 \linewidth]{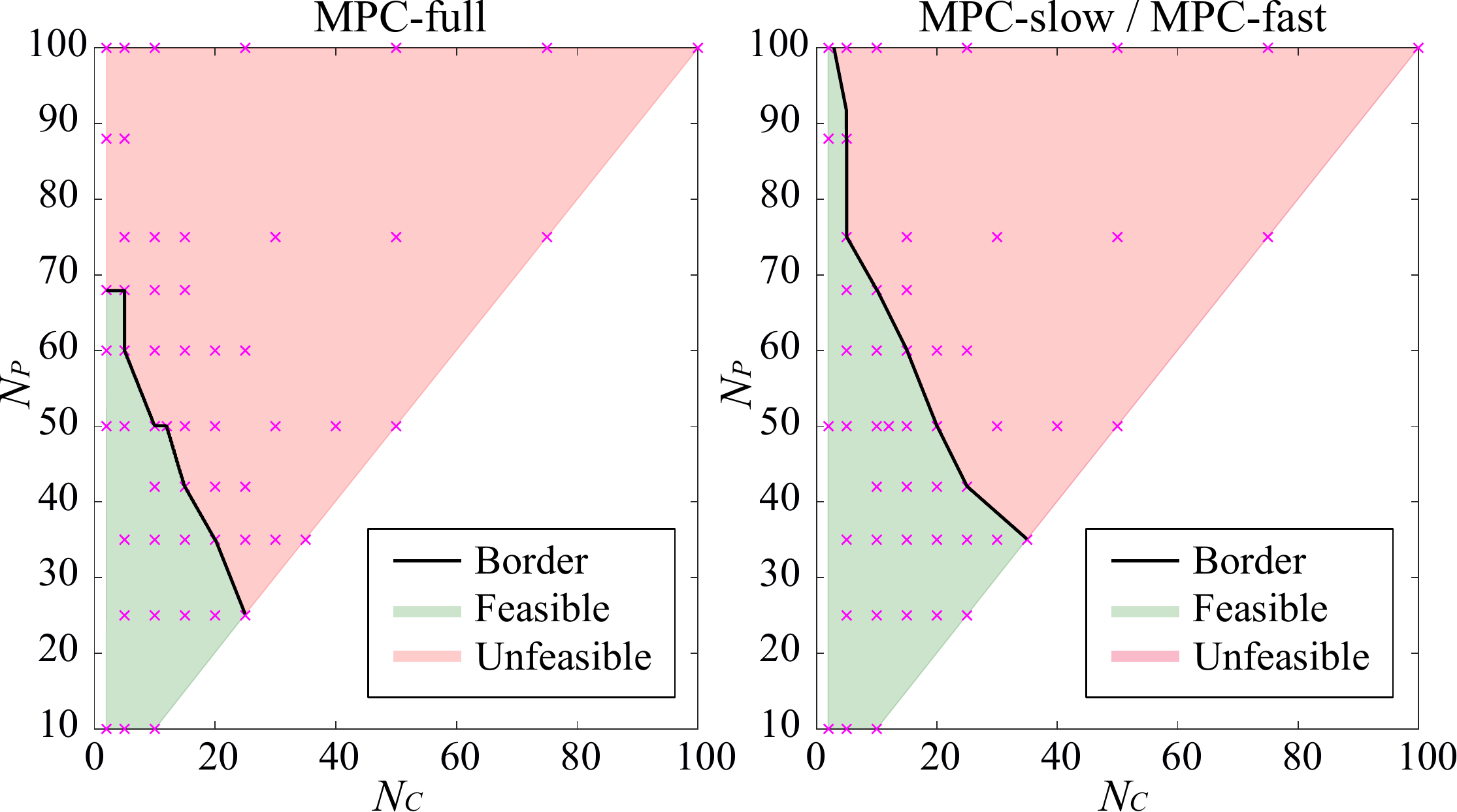}
	\caption{Feasibility check of prediction and control horizon parameters for the different MPC structures. The black line indicates the border of the feasible region to run the controllers in real-time.}
	\label{fig:NC_NP_validity}
\end{figure}
That represents the reference model deployed to predict the link-side dynamical behavior when the gravity effects is compensated.    
\subsection{MPC-Full (motor dynamics \& link dynamics)}
The full model of the flexible-joint robot can be used to obtain the total commanded MPC motor torque directly. In this case the state vector contains both the motor- and the link-side coordinates, rendering \eqref{eq:link_dynamics}--\eqref{eq:motor_dynamics} to
\begin{equation}
\label{mpcfull}
\begin{bmatrix} 
\dq\\ \ddq \\ \dth\\ \ddth  
\end{bmatrix}
\!\!=\!\!
\underbrace{
\begin{bmatrix} 
	\!\vec{0}\!&\!\vec{1}\!&\!\vec{0}\!&\!\vec{0}\!\\\!-\M^{-1}\K\!&\!-\M^{-1}\vec{C}\!&\!\M^{-1}\K\!&\!\vec{0}\!
	\\\!\vec{0}\!&\!\vec{0}\!&\!\vec{0}\!&\!\vec{1}\!\\\!\B^{-1}\K\!&\!\vec{0}\!&\!-\B^{-1}\K\!&\!\vec{0}\!
	\end{bmatrix}}_{\displaystyle \vec{A}_\full}
\!\!\begin{bmatrix} 
\q \\ \dq \\ \thetab \\ \dth
\end{bmatrix}
\!\!+\!\!\underbrace{ 
\begin{bmatrix} 
\!\vec{0}\!\\\!\vec{0}\!\\\!\vec{0}\!\\\!{\B^{-1}}\!
\end{bmatrix}}_{\displaystyle \vec{E}_\full}
\!\vec{\tau}_\text{m}
\end{equation}
The resultant model \eqref{mpcfull} uses twice the number of states compared to the previous ones, namely \eqref{fast_mpc} and \eqref{mpcslow}, which affects the execution time as will be discussed in \sect{sec:simulations_and_experiments}.
\subsection{MPC formulation, objective function}
The MPC formulation presented in the following can be applied to all three model structures, namely \eqref{fast_mpc}--\eqref{mpcfull}.
Therefore, the general formulation of the system dynamics as a function of its states $\vec{z}$ is given:
\begin{align}
\label{eq:SS_dynamics_1}
\dot{\vec{z}} &= \vec{A}_x \vec{z} + \vec{E}_x \vec{u} \\
\label{eq:SS_dynamics_2}
\vec{y} &= \vec{C}_x \vec{z} + \vec{D}_x \vec{u}
\end{align}
Depending on the considered reference model the matrix $\vec{A}_x$ may equal $\vec{A}_\fast$, $\vec{A}_\slow$, or $\vec{A}_\full$. Analogously, the same applies to the matrix $\vec{E}_x$ with respect to $\vec{E}_\fast$, $\vec{E}_\slow$, or $\vec{E}_\full$.
Further, $\vec{C}_x$ is the output matrix, $\vec{D}_x$ is the feed-forward matrix, and $\vec{u}$ is the vector of the control input. 
That leads to different state vectors based on the selected dynamic expression.  
The actuation torque can be determined via the optimization problem
\begin{equation}
\begin{split}
& \min_{\hat{\vec{u}}} (\hat{\vec{y}}-\hat{\vec{y}}_{\mathrm{ref}})^T \vec{Q}_y (\hat{\vec{y}}-\hat{\vec{y}}_{\mathrm{ref}}) + \hat{\vec{u}}^T \vec{Q}_u \hat{\vec{u}}\\
\end{split}
\label{eq:objective_function}
\end{equation}
with
\begin{align}
\vec{Q}_y &= \mathrm{diag}(\vec{Q}_{y,1}, \ldots, \vec{Q}_{y,{N_P}}) \\
\vec{Q}_u &= \mathrm{diag}(\vec{Q}_{u,1}, \ldots, \vec{Q}_{u,{N_C}})
\end{align}
which define the weighting terms for the observed states and the system input, respectively, with ${N_P}$ prediction horizon and ${N_C}$ control horizon, for more details see \cite{van2012subspace}.
Additionally, \eqref{eq:objective_function} is subject to \eqref{eq:SS_dynamics_1}--\eqref{eq:SS_dynamics_2} and the input constraints $\vec{u} \in \vec{\mathcal{U}}$, where $\vec{\mathcal{U}}$ describes the maximum feasible actuator torques.
The predicted output $\hat{\vec{y}} \in\mathbb{R}^{N_P}$ as a function of the current states $\vec{z}_{k}$ and the input vector $\hat{\vec{u}}$ is given by
\begin{equation}
\hat{\vec{y}}
= \hat{\vec{C}}(\vec{A}_x,\vec{C}_x) \vec{z}_{k} + \hat{\vec{D}}(\vec{A}_x,\vec{E}_x,\vec{C}_x,\vec{D}_x) \hat{\vec{u}} \ .
\end{equation}
Further, $\hat{\vec{C}}$ and $\hat{\vec{D}}$ are the augmented discrete output and input matrices.
From an implementation point of view, the objective function is formulated as a quadratic programming problem to find $\hat{\vec{u}}$.
Depending on the selected reference model of the MPC structures the optimization searches the optimal commanded motor torque that complies with \eqref{eq:objective_function} and the imposed constraints.

\section{Simulations and Experiments}\label{sec:simulations_and_experiments}
\begin{figure}[t]
	\centering
	\includegraphics[width=1 \linewidth]{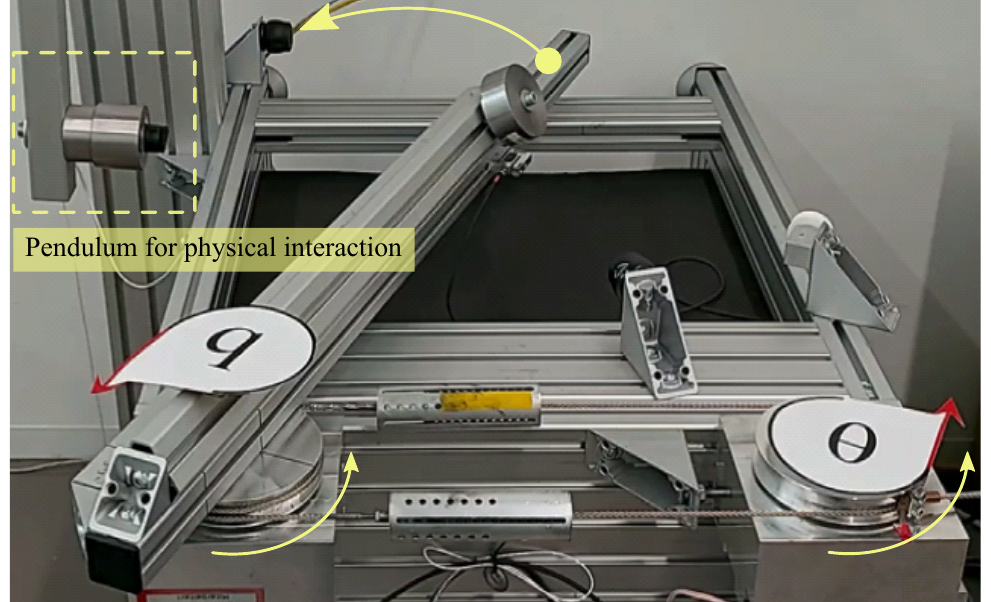}
	\caption{Experimental setup of the flexible joint robot. The system is enhanced with a pendulum-like load to be able to exert external forces.}
	\label{fig:test_bed_SEA}
\end{figure}

The proposed MPC variations are implemented in a cascaded fashion.  
However, both the slow and the fast control loops are running at the same sampling rate in the actual robot implementation.
The experimental validation is conducted using a flexible robot joint constructed by an elastic element from the DLR C-Runner \cite{loeffl2016dlr} with a DLR LWR III drive unit \cite{2007.Schaeffer.DLR_LWR} with a feasible torque range of {$\pm$100\,Nm}, see \fig{fig:test_bed_SEA}. 
The controller gains are reported in Table~\ref{tab:param}, which are represented as the outer-loop control frequency $\omega_{n}$ and damping ratio $\zeta$. Furthermore, the shaping ratio of the apparent motor inertia is $\vec{\gamma}_\mathrm{rf}=\vec{B}\vec{B}_\mathrm{d}^{-1}$ and the damping ratio of the fast torque dynamics is $\zeta_\mathrm{f}$. The parameters of the considered joint are ${B=0.5980\,\mathrm{kgm}^2}$(motor inertia), ${K~=~362}$\,Nm/rad (intrinsic joint stiffness), and ${M~=~1\,\mathrm{kgm}^2}$ for the link inertia.
\begin{figure}[b]
	\centering
	\includegraphics[width=1 \linewidth]{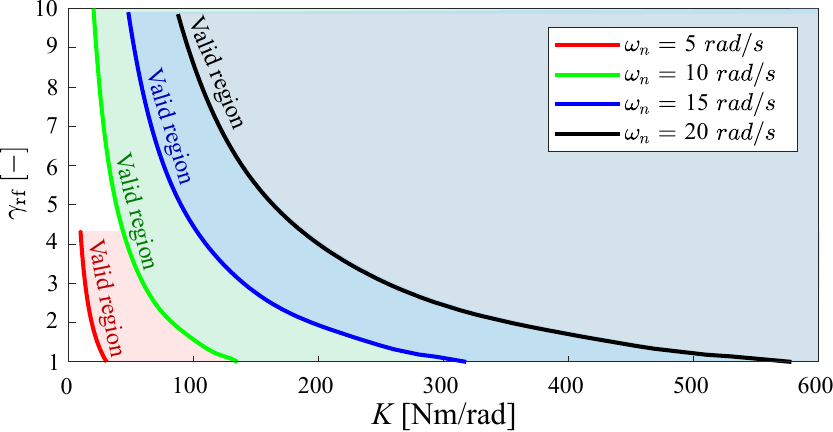}
	\caption{The validity of SP as a function of the joint stiffness and the shaping factor of the motor inertia $\vec{B}\vec{B}_\mathrm{d}^{-1}$. Multiple regions are characterized based on the outer-loop control frequency~$\omega_{n}$~.}
	\label{fig:SP_limits}
\end{figure}
\begin{figure*}
	\centering
	\includegraphics[width=1 \linewidth]{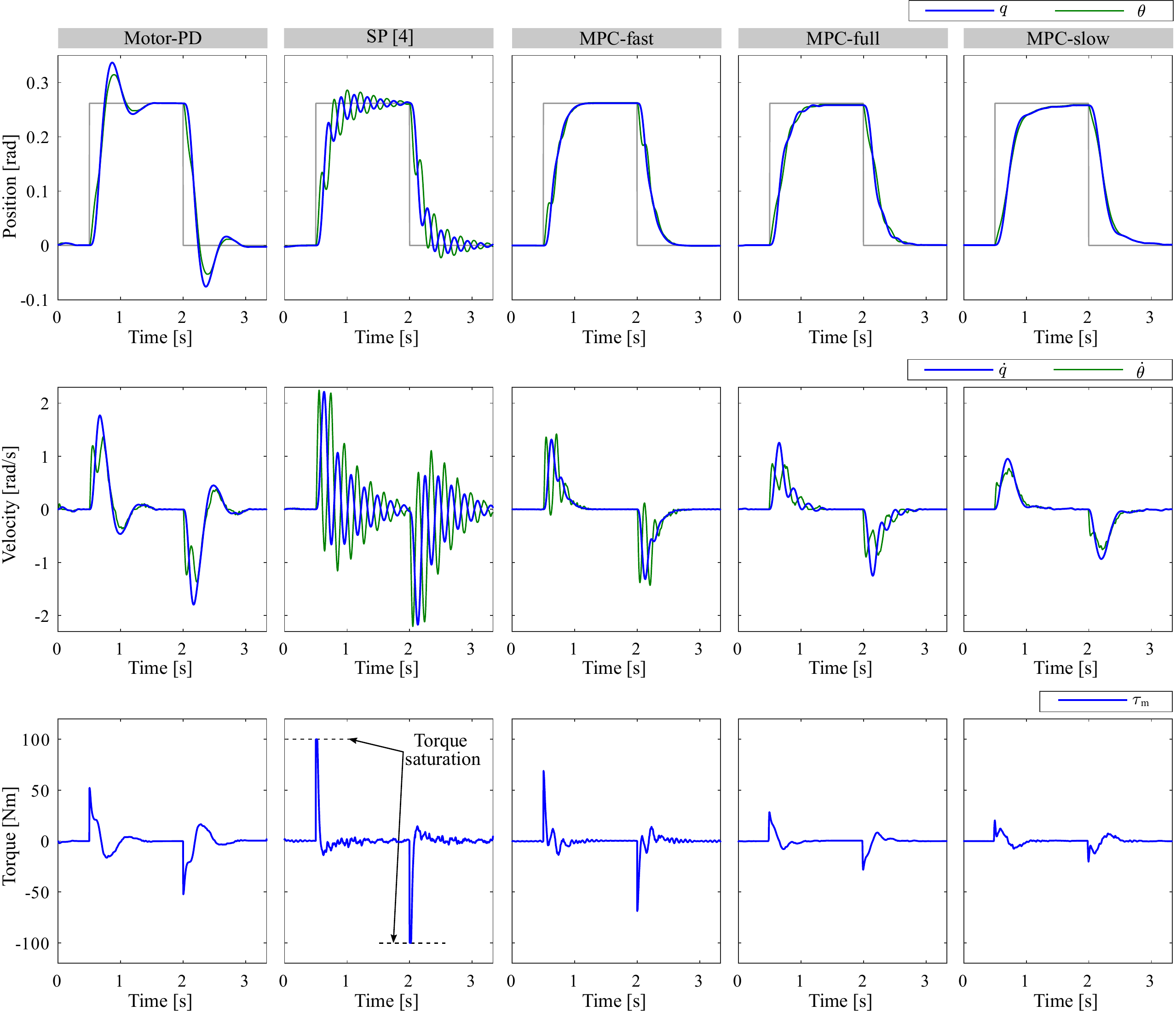}
	\caption{Experimental evaluation of different control techniques on a flexible joint. The performance of the classical motor-PD and singular-perturbation controller are compared with three MPC structures. Link-side and motor-side position and velocity are visualized for each case as well as the associated commanded motor torque.}
	\label{fig:multiplesteps_multipleCtrl}
\end{figure*}
The values for the control and prediction horizons, respectively, are determined experimentally; each of the magenta crosses in \fig{fig:NC_NP_validity} shows a trial/test. The light green area shows the feasible region for $N_P$ and $N_C$, which are larger for MPC-slow and MPC-fast than for MPC-full. This is expected since MPC-full uses twice the number of states and thus involves a numerically more complex model within the prediction. The red borderline shows the maximum allowable region for real-time execution of the controllers. 
\subsection{Validity of the SP method}
The SP theory assumes that there is a sufficiently small parameter responsible for the separation between the fast and the slow dynamics. In the general case of a flexible-joint robot, the parameter $\epsilon$ directly related to the joint stiffness $\vec{K}$ is in control of this separation, see \eqref{eq:sp_K_K0}. 
In \fig{fig:SP_limits}, it can be observed that the validity of the SP assumption depends not only on the joint stiffness but also on the reduction factor of the apparent motor inertia and the outer loop frequency. Practically, this could mean that if the system is not controllable with the SP approach for a specific stiffness, one could increase the reduction factor for the motor inertia or decrease the outer loop control frequency. The solid lines in \fig{fig:SP_limits} represent the region for which the mean average error of the link position over five seconds is larger than~0.26~rad.

\begin{figure}
	\centering
	\includegraphics[width=1 \linewidth]{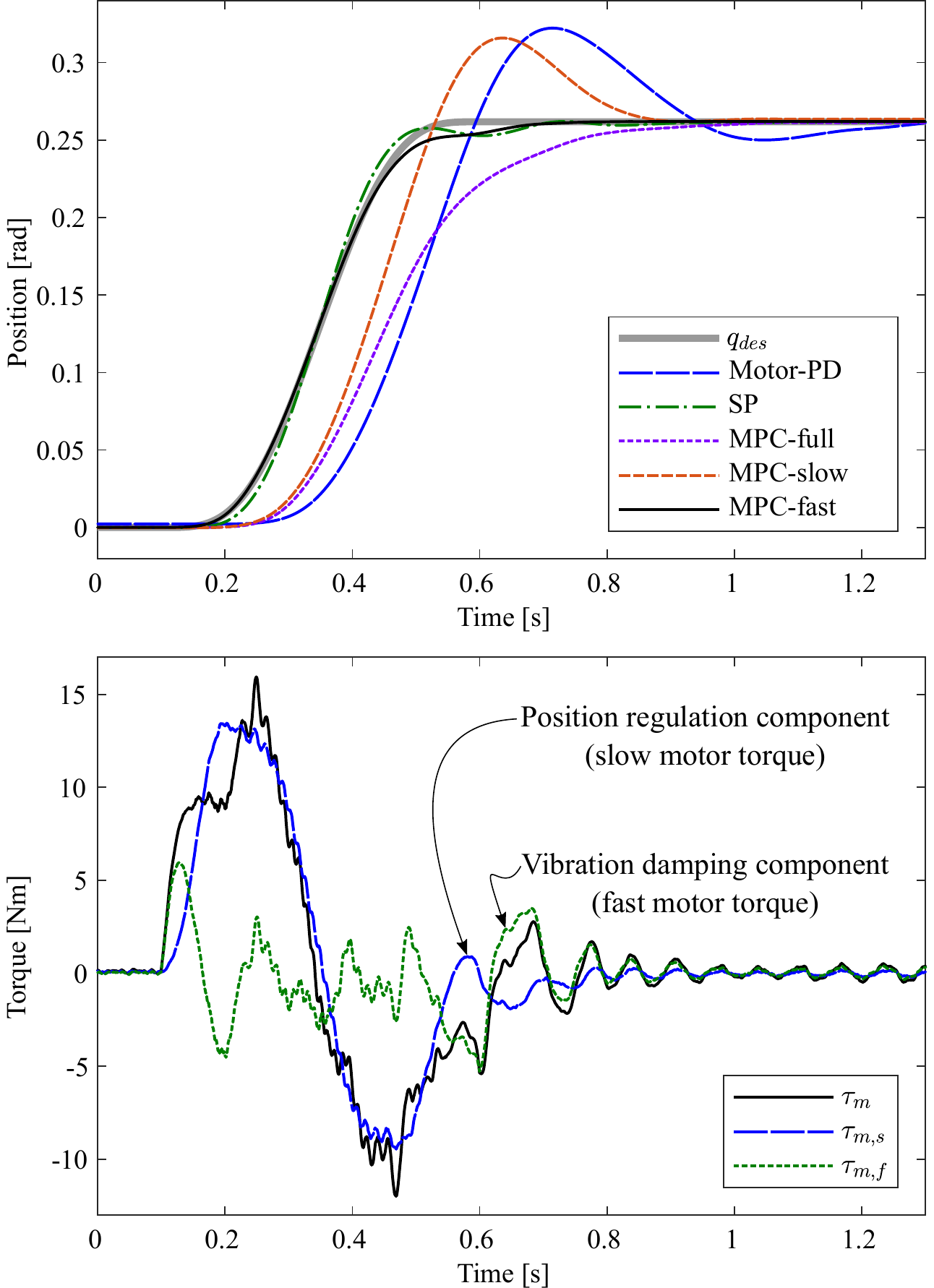}
	\caption{The experimental result of a position trajectory smoothly from 0 to +0.26 rad is shown in the top. In the bottom the total motor torque is shown as two components, one is slow and responsible to achieve the desired position, the other is fast due to the dynamic deflection during motion phase.}
	\label{fig:smooth_step_MPCfast}
\end{figure}
\begin{figure*}
	\centering
	\includegraphics[width=1 \linewidth]{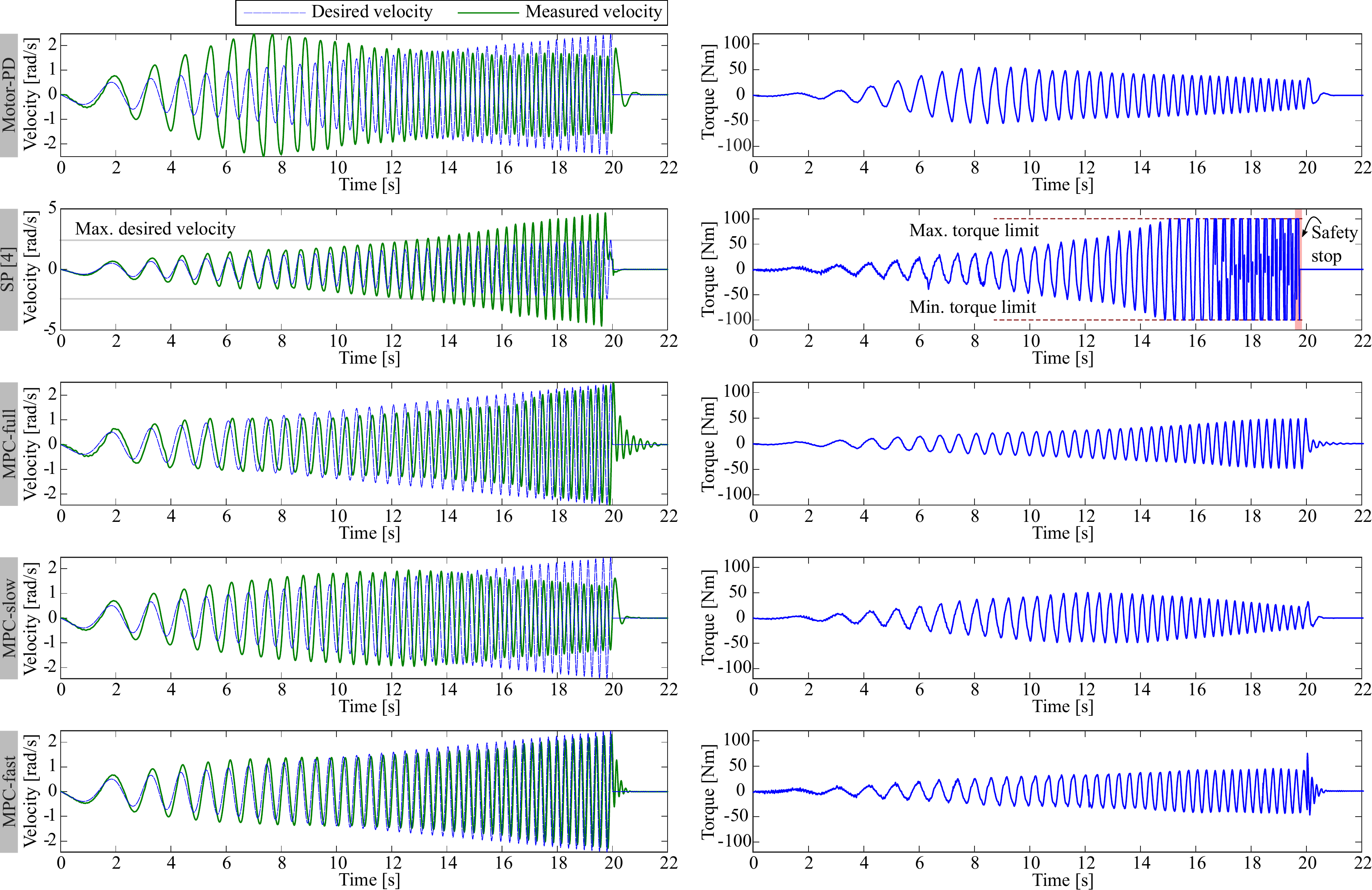}
	\caption{The experimental result when using sinusoidal desired position signal with increasing frequency (chirp signal). The desired and actual link-side velocities are visualized on the left, and on the right, the corresponding motor torque is shown.}
	\label{fig:chirp_MPC_velocity}
\end{figure*}
\begin{table}[h]
	\caption{Parameters of the controllers in the experiments}
	\label{tab:param}
	\begin{tabular}{lccccc}
		\hline
		\textbf{} & $\omega_{n}$, $\zeta$ & $\gamma_\mathrm{rf}$ &$\zeta_\mathrm{f}$& $\vec{Q}_y$                    & $\vec{Q}_u$         \\ \hline
		Motor-PD  & 14 rad/s, 0.7  &     &        &                       &            \\
		SP        & 15 rad/s, 1.0  & 2.0 & 1.0 &                       &            \\
		MPC-full  &                 &     &        & {[}60, 2e-2, 5e-4{]} & {[}2e-6{]} \\
		MPC-slow  &                 & 2.0    & 1.0 & {[}5, 1e-2{]}         & {[}1e-5{]} \\
		MPC-fast  & 15 rad/s, 1.0  &     &        & {[}1, 5e-3{]}         & {[}1.3{]}  \\ \hline
	\end{tabular}
\end{table}
\subsection{Link-side position and impedance control}
The performance of the position and impedance control is validated for all MPC variations together with the standard motor-PD controller and the standard SP implementation \cite{2008.Ott.CartImp}. 
The parameters have been tuned to achieve comparable rising times for all controllers to provide conditions for fair comparisons. All controllers (except the motor-PD) are applied to the link dynamics and actively use $(\q,\dq)$ as feedback signals. Figure \ref{fig:multiplesteps_multipleCtrl} shows the experimental results for the step responses, the associated velocities, and generated motor torques. Notably, overshoot can be observed in the case of the classical motor-PD controller as it ignores the flexibility in the joint. The SP controller \eqref{eq:SP_control} shows continuous oscillations as well as commanded torque saturation. Figure \ref{fig:multiplesteps_multipleCtrl} shows that the MPC versions minimize the generated motor torque as a result of the respective objective functions. Interestingly, the proposed MPC-fast structure shows smooth and faster convergence in the link-side position. With MPC-slow and SP in the inner control loop, slower convergence can be observed due to the limited performance of the classical SP.      
The MPC-full approach is expected to achieve superior performance as it uses the complete model in the prediction phase, but due to the limitation of increasing the value of the prediction horizon $N_P$, the performance is limited in the real implementation.

Additionally, a smooth trajectory with a step from 0 to 0.26 rad is applied to evaluate the tracking performance generated through a septic polynomial to ensure that it is at least two times differentiable. In \fig{fig:smooth_step_MPCfast} the different MPC structures are compared together with the standard SP approach and the motor-PD controller. The MPC-fast variant shows the lowest link position error. To gain a closer look of the control action, the individual components of the commanded motor torque are also visualized in \fig{fig:smooth_step_MPCfast}\,(bottom) for the MPC-fast case. It can be seen that the total motor torque consists of two components, one is slow and responsible for position regulation/tracking, the other is relatively fast and actively damps oscillations introduced by the joint elasticity. This result reflects the physical interpretation of the control action, which also incorporates the future prediction of the fast dynamics.    
As the trajectory here is less demanding, the standard SP controller behaves better \wrt the step command case. However, undesirable oscillations can be noticed while following a more dynamical trajectory.        

Further, a chirp signal (from 0 to approx. 4Hz) is used as a desired position to validate the system for more dynamical effects. The Root Mean Square Error (RMSE) for the position and velocity are depicted in Table~\ref{tab:RMSE}, and in \fig{fig:chirp_MPC_velocity} the velocities and the corresponding motor torques are shown. The motor-PD controller features an early velocity deviation compared to the link-side controllers. The SP approach shows a noticeable high velocity amplitude (up to 4.7 rad/s) when the desired signal frequency increases as the commanded motor torque has already reached the saturation limits. In this case, the experiment is terminated through the activation of the safety stop and before the end of the trajectory execution at 20 sec, see \fig{fig:chirp_MPC_velocity}. Here, MPC-full and MPC-slow controllers have a comparatively high RMSE over the whole range of motion due to phase shift \wrt the desired signals. The MPC-fast controller has a comparable position error to the SP approach but a smaller velocity error. It also can be seen from \fig{fig:chirp_MPC_velocity} that all MPC variations do not exceed the actuation torque limits when the frequency of the desired trajectory is increased.
\begin{table}[b]
	\caption{Root Mean Square Error (RMSE)}
	\label{tab:RMSE}
	\begin{tabular}{lccc}
		\hline
		\textbf{} & Pos. RMSE [rad] & Vel. RMSE [rad/s] & Additional info.        \\ \hline
		Motor-PD  & 0.1945   & 2.1071   & --          \\
		SP        & 0.0893   & 1.4248   & Safety stop \\
		MPC-full  & 0.1364   & 1.5461   & --          \\
		MPC-slow  & 0.1509   & 1.7658   & --          \\
		MPC-fast  & 0.0803   & 0.7692   & --          \\ \hline
	\end{tabular}
\end{table}
\begin{figure}[t]
	\centering
	\includegraphics[width=1 \linewidth]{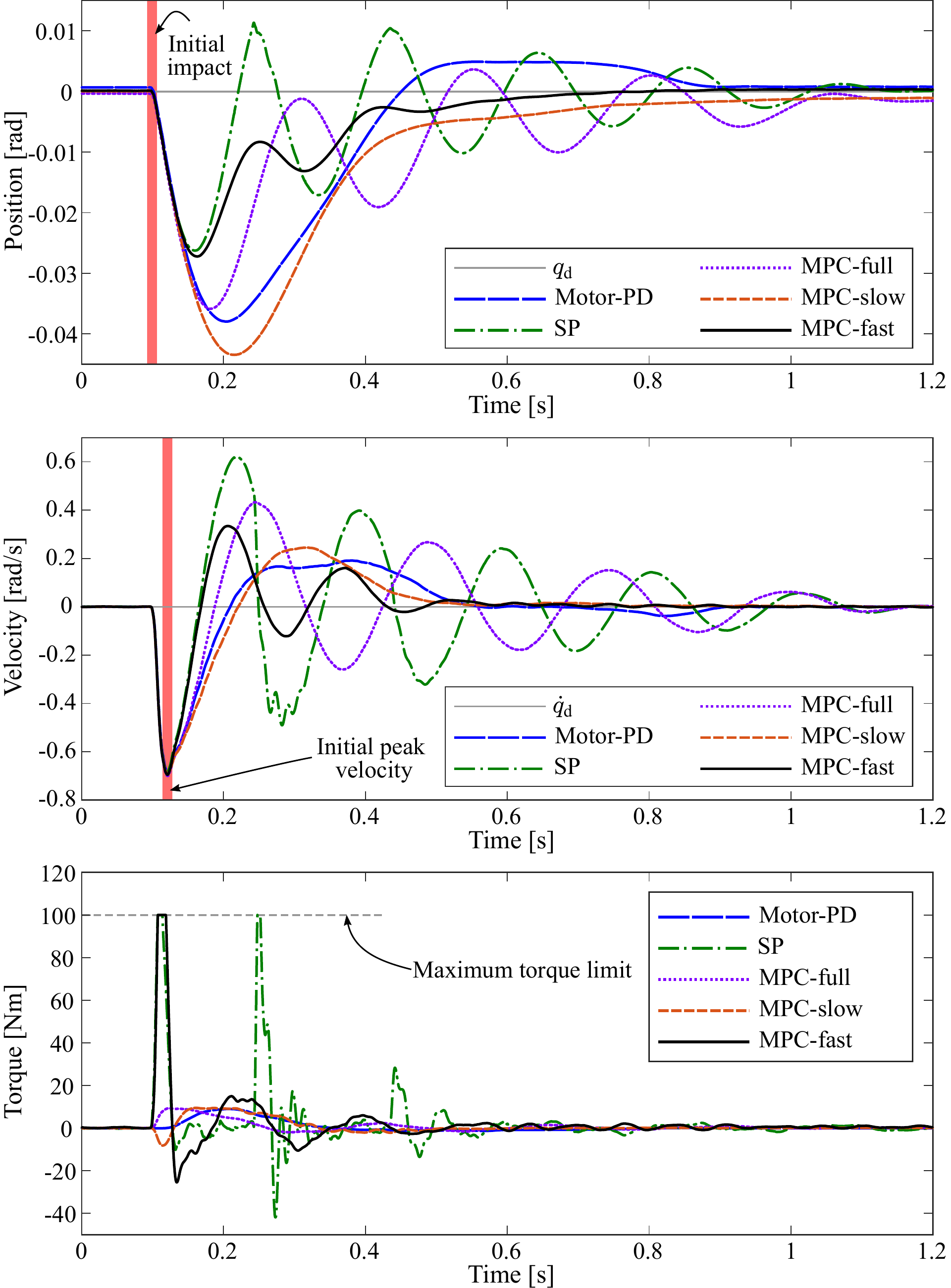}
	\caption{The experimental evaluation of the physical interaction behavior. The system is excited using the same impact force through a pendulum-like external load.}
	\label{fig:interaction_MPC_multicont}
\end{figure}
\subsection{Physical interaction behavior}
The interaction behavior is illustrated by means of an external force applied through a pendulum-like load, see \fig{fig:test_bed_SEA}.  
In the experiments, the system is excited with the same amount of energy by shifting the pendulum to a fixed distance from the link. The injected energy can be seen by the comparable initial peak velocity for the considered controllers, see \fig{fig:interaction_MPC_multicont} (middle). Interestingly, the SP approach saturates two times unnecessarily by the effect of the initial impact. The MPC-fast controller features superior vibration/oscillation damping as it aims to control the torque dynamics. 
From a human-robot interaction perspective, the MPC-fast controller is more transparent to parameterize in order to achieve the desired interaction behavior. This is because it can be interfaced with a link-side impedance controller as an outer control loop such that the desired stiffness can be specified accurately. Simultaneously, the torque tracking performance is maintained in the inner control loop through the MPC while respecting the torque limits.
\section{Conclusions}\label{sec:conclusions}
A method to derive a joint-level controller for flexible joint robots based on MPC techniques was presented. Motivated by the two-time-scales separation from the singular perturbation theory, three different MPC design structures were investigated. The first one uses the fast torque dynamics as a reference model and aims at regulating the torque control loop within the joint torque limits. The second one considers the slow link-side dynamics and is used to design an outer position/impedance control loop. The third one exploits the full model the flexible-joint robots to generate the commanded torque respecting the actuator constraints. The proposed methods have been tested experimentally on a flexible-joint setup.  
Compared to the state of the art, significant improvements in the dynamic link-side position errors were achieved. 
This work also provides more insights into the validity of the singular-perturbation assumption for flexible-joint robot dynamics.

\section{Acknowledgment}\label{sec:Acknowledgement}
The authors would like to thank Martin Görner for the drawing in \fig{fig:intro_fig} and Florian Loeffl for his support with hardware.


\balance
\bibliographystyle{IEEEtran}
\bibliography{references}

\end{document}